# NATURAL LANGUAGE PROCESSING IN BIOMEDICINE: A UNIFIED SYSTEM ARCHITECTURE OVERVIEW

**Son Doan[1], Mike Conway[2], Tu Minh Phuong[3], Lucila Ohno-Machado[1]**

[1]Division of Biomedical Informatics, University of California, San Diego

9500 Gilman Dr, La Jolla, CA 92093

[2]Division of Behavioral Medicine, University of California, San Diego

9500 Gilman Dr, La Jolla, CA 92093

[3]Department of Computer Science, Posts and Telecommunications Institute of Technology, Hanoi, Vietnam

In modern electronic medical records (EMR) much of the clinically important data – signs and symptoms, symptom severity, disease status, etc. – are not provided in structured data fields, but rather are encoded in clinician generated narrative text. Natural language processing (NLP) provides a means of "unlocking" this important data source for applications in clinical decision support, quality assurance, and public health. This chapter provides an overview of representative NLP systems in biomedicine based on a unified architectural view. A general architecture in an NLP system consists of two main components: *background knowledge* that includes biomedical knowledge resources and *a framework* that integrates NLP tools to process text. Systems differ in both components, which we will review briefly. Additionally, challenges facing current research efforts in biomedical NLP include the paucity of large, publicly available annotated corpora, although initiatives that facilitate data sharing, system evaluation, and collaborative work between researchers in clinical NLP are starting to emerge.

## Introduction

In modern electronic medical records (EMR) most of the clinically important data – signs and symptoms, symptom severity, disease status, etc. – is not provided in structured data fields, but are rather encoded in clinician-generated narrative text. Natural language processing (NLP) provides a means of "unlocking" this important data source, converting unstructured text to structured, actionable data for use in applications for clinical decision support, quality assurance, and public health surveillance. There are currently many NLP systems that have been successfully applied to biomedical text. It is not our goal to review all of them in this chapter, but rather to provide an overview of how the field evolved from producing monolithic software built on platforms that were available at the time they were developed to contemporary component-based systems built on top of general frameworks. More importantly, the performance of these systems is tightly associated with their "ingredients" (i.e., modules that are used to form its background knowledge), and how these modules are combined on top of the general framework. We highlight certain systems based on their landmark status as well as on the diversity of components and frameworks they are based on.

The Linguistic String Project (LSP) was an early project starting in 1965 that focused on medical language processing [1]. The project created a new schema for representing clinical text and a dictionary of medical terms in addition to addressing several key clinical NLP problems such as de-identification, parsing, mapping, and normalization. The system's methodology and architecture have substantially influenced many subsequent clinical NLP systems.

One of the main requirements for developing clinical NLP systems is a suitable biomedical knowledge resource. The Unified Medical Language System (UMLS) [2], initiated in 1986 by National Library of Medicine, is the most widely used knowledge resource in clinical NLP. The UMLS contains controlled vocabularies of biomedical concepts and provides mappings across those vocabularies.

With the development of machine learning, NLP techniques, and open-source software, tools have been developed and are now available in open source (e.g., NLTK[1], Mallet[2], Lingpipe[3], and OpenNLP[4]). These tools can help biomedical researchers re-use and adapt NLP tools efficiently in biomedicine. Several software frameworks that facilitate the integration of different tools into a single pipeline have been developed, such as GATE[5] (General Architecture for Text Engineering) and UIMA[6] (Unstructured Information Management Architecture). Given the success of IBM's Watson in the 2011 Jeopardy challenge, the UIMA framework, which was used for real-time content analysis in Watson, has now been applied widely by the biomedical NLP community. The highly recognized open source system cTAKES was the first clinical NLP system to use the UIMA framework to integrate NLP components and is rapidly evolving.

---

[1] http://www.nltk.org

[2] http://mallet.cs.umass.edu/

[3] http://alias-i.com/lingpipe/

[4] http://opennlp.apache.org/

[5] http://gate.ac.uk/

[6] http://uima.apache.org/

In this chapter, we provide an overview of NLP systems from a unified perspective focused on system architecture. There are already comprehensive reviews and tutorials about NLP in biomedicine. Spyns provided an overview of pre-1996 biomedical NLP systems [3], while Demner-Fushman et al. more recently reviewed and summarized NLP methods and systems for clinical decision support [4]. The use of NLP in medicine has been comprehensively reviewed by Friedman [5], Nadkami et al. [6], and more recently by Friedman and Elhadad [7]. The review in this chapter differs from previous work in that it emphasizes the historical development of landmark clinical NLP systems, and presents each system in light of a unified system architecture.

We consider that each NLP system in biomedicine contains two main components: biomedical background knowledge and a framework that integrates NLP tools. In the rest of this paper, we will first outline our model architecture for NLP systems in biomedicine, before going on to review and summarize representative NLP systems, starting with an early NLP system, LSP-MLP, and closing our discussion with the presentation of a more recent system, cTAKES. Finally, we will discuss challenges as well as trends in the development of current and future biomedical NLP systems.

## Materials

### *A general architecture of an NLP system in biomedicine*

We start from a discussion by Friedman and Elhadad [8] in which NLP and its various components are illustrated, as reproduced in Fig. 1. NLP aspects can be classified into two parts in the figure: the left part contains trained corpora, domain model, domain knowledge, and linguistic knowledge; the right part contains methods, tools, systems, and applications. From the viewpoint of system architecture, we consider a general architecture in which an NLP system contains two main components: *background knowledge*, which corresponds to the left part of the figure, and a *framework* that integrates NLP tools and modules, which corresponds to the right part of the figure. Our view of a general architecture is depicted in Fig. 2. Below we describe the two main components and their roles in biomedical NLP systems.

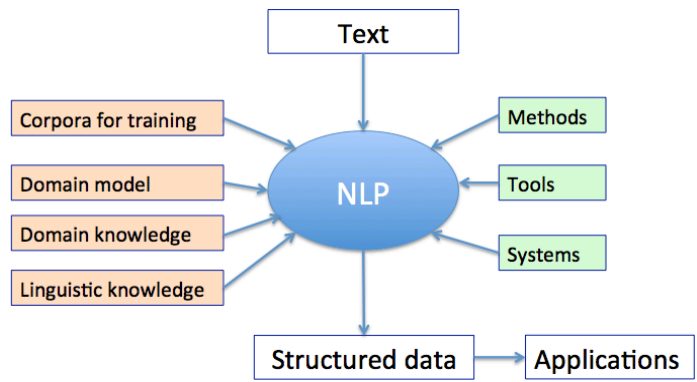

**Fig. 1.** Aspects of clinical NLP systems as described by Friedman and Elhadad [8]. The rectangles on the left side represent background knowledge, and the components on the right side represent the framework (i.e., algorithms and tools). Background knowledge and framework are the main components of an NLP system.

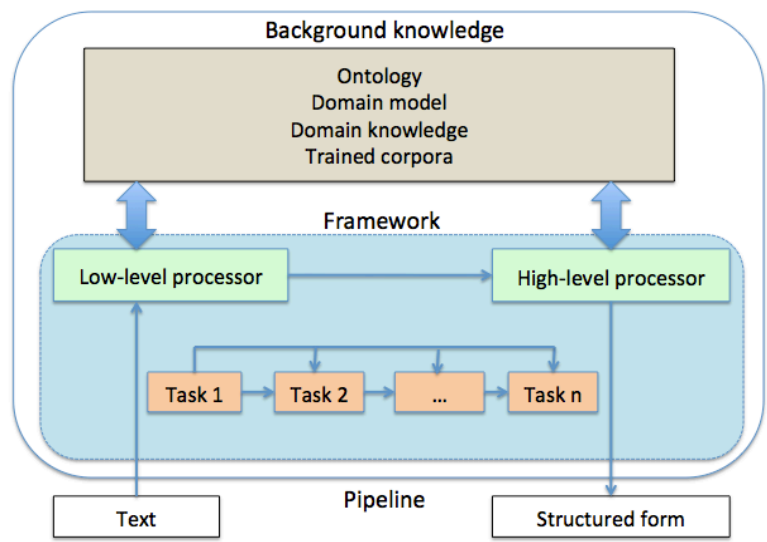

**Fig. 2.** A general architecture of a clinical NLP system contains two main components: background knowledge and framework. Background contains ontologies, a domain model, domain knowledge, and trained corpora. Framework includes a low-level processor for tasks such as tokenization and part-of-speech tagging. A high-level processor is used for tasks such as named entity recognition and relation extraction. Tasks or modules in the framework can be dependent or independent and are organized sequentially or hierarchically.

*Background knowledge for NLP in biomedicine: The Unified Medical Language System (UMLS)*

As mentioned in the introduction, biomedical knowledge is an important component in building clinical NLP systems. Domain knowledge and linguistic knowledge are key elements. Earlier systems such as LSP-MLP built their own medical vocabulary and tools due to the lack of easily available resources at that time. The creation of the Unified Medical Language System (UMLS), which began development in 1986, substantially benefited clinical NLP systems. The UMLS contains three main components: the Metathesarus, the Semantic Network, and the SPECIALIST lexicon. For practical purposes, the UMLS can be considered to be like an ontology of biomedical concepts and their relations. Below we briefly summarize each component of the UMLS.

The UMLS's *Metathesarus* currently contains over 1 million biomedical concepts and 5 millions concept names originating from over 150 controlled vocabularies in the biomedical sciences, such as ICD-10, MeSH, SNOMED CT, and RxNorm.

The UMLS *Semantic Network* provides a consistent categorization of all concepts represented in the UMLS Metathesaurus. It reduces the complexity of the Metathesaurus by grouping concepts according to semantic types. Currently it contains 135 broad categories and 54 relationships among categories. For example, the category "Disease or Syndrome" has a relationship "associated_with" with the category "Finding", and the category "Hormone" has a relationship "Affects" with the category "Disease or Syndrome" in the semantic network.

The UMLS *SPECIALIST lexicon* contains syntactic, morphological, and spelling information for biomedical terms [9]. Currently it contains over 200,000 terms and is used by the UMLS lexical tools for NLP tasks.

Background knowledge also includes domain models and trained corpora, which are used to deal with specific domains such as radiology reports, pathology reports, and discharge summaries. Annotated corpora are manually marked-up by human annotators and used to train machine learning linguistic classifiers, as well as to evaluate rule-based systems.

*NLP tools and integrated frameworks*

There are two main approaches for building NLP tools. The first is rule-based, which mainly uses dictionary look-up and rules. The second uses a machine learning approach that relies on annotated corpora to train learning algorithms. Early systems often used rule-based approach since they were relatively easy to design and implement. Currently, with the development of robust statistical machine learning methods and an increasing number of annotated corpora, many clinical NLP systems have moved away from relying exclusively on rule-based methods, although there is still a high cost in generating new annotated training data, which are still required to account for differences in tasks, types of documents, as well as their provenance. As shown in many clinical NLP challenges, machine learning methods often achieve better results than rule-based methods. However, rule-based methods are somewhat easier to customize and adapt to a new domain. Most contemporary NLP systems are hybrid, i.e., built from a combination of rule-based and machine learning methods [8].

Fig. 2 shows how NLP tools can be integrated into a pipeline built on top of a particular framework. By framework we mean a software platform for the control and management of components such as loading, unloading, and handling components of the pipeline. Components within a framework can be embedded and linked together or used as plug-ins. For NLP systems in biomedicine, the framework can be divided into two levels: low-level and high-level processors. *Low-level processors* perform foundational tasks in NLP such as sentence boundary detection, section tagging, part-of-speech tagging, and noun phrase chunking. *High-level processors* perform semantic level processing such as named entity recognition (e.g., diseases/disorders, sign/symptoms, medications), relation extraction, and timeline extraction.

The framework can be integrated into the NLP system itself or it can leverage available general architectures. The two most widely used general architectures are GATE[7] (General Architecture for Text Engineering) and UIMA[8] (Unstructured Information Management Architecture). Both consist of open-source software.

GATE, written in Java, was originally developed at the University of Sheffield in 1995 and is widely used in the NLP community. It includes basic NLP tools for low-level processing (such as tokenizers, sentence splitters, part-of-speech taggers) packaged in a wrapper called CREOLE, and a high-level processor for named entity recognition packaged in an information extraction system called ANNIE. It can integrate available NLP tools and machine learning software such as Weka[9], RASP[10], SVM Light[11], and LIBSVM[12]. Several clinical NLP systems have used GATE as their framework, such as HITEx (which will be in the next section), and caTIES[13] for cancer text information extraction.

---

[7] http://gate.ac.uk/

[8] http://uima.apache.org/

[9] http://www.cs.waikato.ac.nz/ml/weka/

[10] http://www.sussex.ac.uk/Users/johnca/rasp/

[11] http://svmlight.joachims.org/

[12] http://www.csie.ntu.edu.tw/~cjlin/libsvm/

[13] http://caties.cabig.upmc.edu/

UIMA, written in Java/C++, was original developed by IBM and is part of the Apache Software Foundation software since 2006. Its motivation is "to foster reuse of analysis components and to reduce duplication of analysis development. The pluggable architecture of UIMA allows to easily plug-in your own analysis components and combine them together with others." [14] The framework is best known as the foundation of IBM's 2011 Jeopardy challenge Watson system. UIMA's functionalities are similar to GATE but are more general since UIMA can be used for analysis of audio and video data, in addition to text. There are several clinical NLP systems that use the UIMA framework such as cTAKES (described in the next section), MedKAT/P[15] for extracting cancer-specific characteristics from text, and MedEx [10, 11] (Java version)[16] for medication extraction.

*System selection*

In order to give a unified view of system architecture, we selected representative NLP systems in this review. We selected systems due to their historical importance and influence in the biomedical NLP community.

We first chose two widely influential landmark clinical NLP systems: LSP-MLP and MedLEE. LSP-MLP is a pioneering project and has greatly influenced subsequent NLP systems, and MedLEE is a system that is currently widely used in clinical NLP communities. We then selected a specific-purpose system called SymText, which was designed for radiology report processing. SymTex began development in the 1990s and is still in active use today. We also briefly review MetaMap, a widely used tool in the biomedical NLP community. We chose two systems based on GATE and UIMA: HITEx, and cTAKES, respectively. Summaries of characteristic features of the clinical NLP systems reviewed in this chapter are presented in Table 1.

---

[14] http://uima.apache.org/doc-uima-why.html
[15] http://ohnlp.sourceforge.net/MedKATp/
[16] http://code.google.com/p/medex-uima/

**Table 1.** Summary of characteristic features of some representative clinical NLP systems.

| System | Programming language | Creator | Framework | Open/Closed source and License | Background knowledge resource | Clinical domain or source of information | Encoding |
|---|---|---|---|---|---|---|---|
| LSP-MLP | Fortran C++ | New York University | | Software provided by Medical Language Processing LLC corporation | Developed its own medical lexicons and terminologies | Progress note, clinical note, X-ray report, discharge summary | SNOMED |
| MedLEE | Prolog | Columbia University | | Closed source Commercialized by Columbia University and Health Fidelity, Inc. | Developed its own medical lexicons (MED) and terminologies | Radiology Mammography, discharge summary | UMLS's CUI |
| SPRUS/ SymText/ MPLUS | LISP, C++ | University of Utah | | Closed source | UMLS | Radiology Concepts from findings in radiology reports | ICD-9 |
| MetaMap | Perl, C, Java, Prolog | National Library of Medicine | | Not open source but free available under UMLS Metathesaurus License Agreement | UMLS | Biomedical text Candidate and mapping concepts from UMLS | UMLS's CUI |
| HITEx | Java | Harvard University | GATE | Open source i2b2 software license | UMLS | Clinical narrative Family history concept, temporal concepts, smoking status, principal diagnosis, co-morbidity, negation | UMLS's CUI |
| cTAKES | Java | Mayo clinic and IBM | UIMA | Open source Apache 2.0 | UMLS + Trained models | Discharge summary, clinical note Clinical named entities (diseases/disorders, signs/symptoms, anatomical sites, procedures, medications), relation, co-reference, smoking status classifier, side effect annotator | UMLS's CUI and RxNorm |

# Methods

*Systems*

*Linguistic String Project (LSP) – Medical Language Processor (MLP)*

The Linguistic String Project (LSP)[17] was developed in 1965 at New York University by Sager et al. [1, 12]. It is one of the earliest research and development project in computer processing of natural language. The development of LSP was based on the linguistic theory of Zellig Harris: linguistic string theory, transformation analysis, and sublanguage grammar [13–15]. It mainly focused on medical language processing, including the sublanguage of clinical reporting, radiograph reports, and hospital discharge summaries. The LSP approach used a parsing program to identify the syntactic relations among words in a sentence. The project strongly influenced subsequent clinical NLP projects. The LSP's system was called the Medical Language Processor (MLP).

The core component of MLP is a parser. The authors first developed a general NLP parser for the general English language domain, including English grammar and lexicon, and then they extended the system to the sublanguage of biomedicine by adding a medical lexicon and corresponding grammar. Below we summarize the main components of MLP.

> *Background knowledge*
>
> - Lexicons: MLP developed lexicons for both general English language and medical knowledge. In the lexicon, each word has an associated part-of-speech and grammatical and medical "attributes" called subclasses. The lexicon has 60 possible verb objects and 50 medical sub-classes. It also had lists of pre-defined prepositions, abbreviations, and doses. These attributes are used throughout the processing to guide the parsing and to resolve ambiguities. Pre-defined lists consist of
>     - Standard numbers, times and dates
>     - Medical terms
>     - Dose strings
>     - Organism terms
>     - Geographic nouns
>     - Patient nouns
>     - Institution/ward/service nouns
>     - Physician/staff nouns

---

[17] http://www.cs.nyu.edu/cs/projects/lsp/

- Grammar: The grammar is written in Backus-Naur Form (BNF). It finds grammatical structures in clinical text and contains following components:
    - BNF: the context-free component
    - The RESTR (restriction) contains procedures written in the MLP's "Restriction Language". Those procedures test the parse tree for the presence or absence of particular features
    - The LISTS contains lists used in procedures other than RESTR

*Pipeline*

- The *Preprocessor* breaks input text into sentences. Then, the preprocessor identifies possible spelling errors, abbreviations, all forms of names of patients, staff, facilities, and administrative and geographic areas for de-identification. Numbers, units, and dates are transformed into ANSI standard format.
- The *MLP Parser* uses a top-down, context-free grammar-based parser. The system generates multiple parses of ambiguous sentences guided by a BNF grammar. The parser was originally written in FORTRAN, and then partly converted into Prolog [16]. Today it is written in C++. The MLP system is now publicly available through the Web site provided by Medical Language Processing, LLC - a Colorado corporation[18].

The parser proceeds from left to right through the sentence, and top to bottom through the BNF definitions. Once the parser associates a terminal symbol of the parse tree, the attributes of the word can be tested by a restriction, for example, the agreement of subject and verb. The following steps are involved in the processing of text:

- *Selection* passes or rejects a parse based on subtrees.
- *Transformation* decomposes sentences into their basic canonical sentences.
- *Regularization* connects basic canonical sentences by conjunctions.
- *Information format* maps the syntactic parse trees into medical information structures. MLP considers 11 information structures related to patients such as patients, family, medication, treatments, lab/test, etc.

Finally, the output is written into two formats: tab delimited and XML format.

---

[18] http://mlp-xml.sourceforge.net/

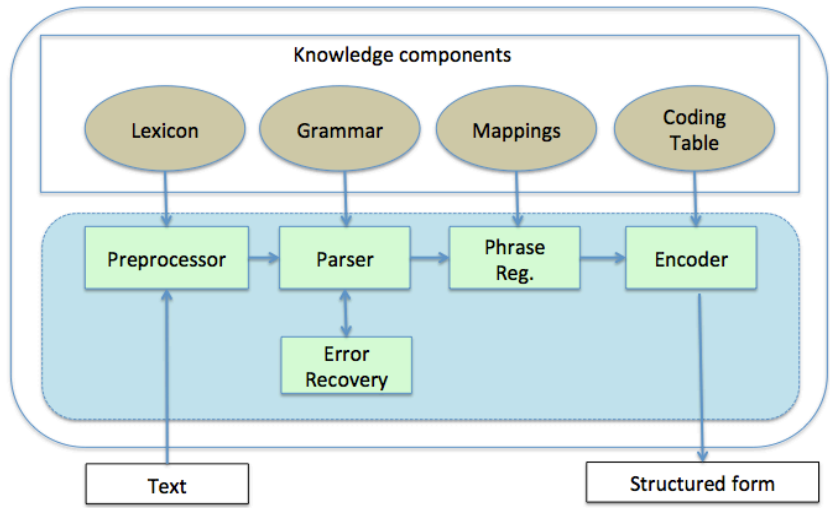

**Fig. 3.** Architecture of MedLEE, where the background knowledge contains components for the lexicon, grammar, mappings, and coding tables. The low-level processor is a preprocessor and the high-level processor consists of modules for parsing, error recovery, phrase regularization, and encoding.

LSP-MLP was used for processing clinical narratives in English, and it was also extended into other languages such as French, German, and Dutch [1]. It has been used to map clinical text into SNOMED codes [17, 18]. LSP-MLP was designed for information retrieval from clinical text, hence there were no reports evaluating mapping. The performance in information retrieval tasks indicated 92.5% recall and 98.6% precision [18]. With its complete structures, LSP-MLP provided an early successful example for the development of subsequent NLP systems.

*MedLEE*

The MedLEE (Medical Language Extraction and Encoding System) system was developed by Friedman et al. at Columbia University [19, 20] in 1994. It was first designed for radiology reports, and was then extended to other domains such as discharge summaries. The system was written in Quintus Prolog. MedLEE contains two main components: (1) a knowledge base including medical concepts, and (2) a natural language processor. MedLee was the first NLP system used as part of a system for actual patient care and some systems in which it was embedded have been shown to improve care [21, 22]. It was commercialized in 2008. The architecture of MedLEE is depicted in Fig. 3.

*Background knowledge*

MedLEE's knowledge base is called the Medical Entities Dictionary (MED) [20], which contains a knowledge base of medical concepts and their taxonomic and semantic relations. Each concept in MED is assigned to an identifier. The MED originally contained over 34,000 concepts.

*Pipeline*

The natural language processor has three phases of processing as follows.

- Phase 1: Parsing. Identify the structures of the text through use of a grammar. It contains three main components: a set of grammar rules, semantic patterns, and lexicon.
  - Grammar rules: MedLEE uses a BNF grammar, originally contained 350 grammar rules.
  - Semantic classes. MedLEE considers sentences contain semantic patterns connected by conjunctions. Semantic patterns can be a word, phrase and/or belong to a semantic class. Examples of semantic classes are Bodyloc, Cfinding, and Disease. MedLEE also considers negation as a semantic pattern in its grammar.
  - Lexicon. The semantic lexicon originally contains both single words (1,700) and phrases (1,400).

- Phase 2: Phrase regularization. This module regularizes the output forms of phrases that are not contiguous. This is a critical step that further reduces the variety that occurs in natural language. The method is automatically applied by processing all phrasal lexical entries that begin with the symbol phrase. Phrase is used to specify that a phrase may occur in a non-contiguous variant form.

- Phase 3: Encoding. This step maps the regularized structured forms to controlled vocabulary concepts. This process is accomplished using a knowledge base containing synonymous terms. The synonym knowledge base consists of associations between standard output forms and a controlled vocabulary. At the end of this stage of processing, the only values that remain are unique controlled vocabulary concepts.

The output of MedLEE is represented as a formal model of clinical information in the domain of interest such as radiology. It has been extended to map extracted concepts into UMLS codes [23], and its architecture was also extended to build an information extraction system for molecular pathways from journal articles [24]. Evaluation on 150 randomly sentences from clinical documents achieved 0.77

Recall and 0.83 Precision compared to 0.69-0.91 Recall and 0.61-0.91 Precision for seven domain experts performing the same tasks [23].

*SPRUS/SymText/MPLUS*

SPRUS/SymText/MPLUS [25–28] was developed in 1994 by Haug et al. at the University of Utah. It has been implemented using common LISP, the Common Lisp Object System (CLOS), and C++. The original system was called SPRUS, and it evolved into SymText (Symbolic Text Processor), NLUS (Natural Language Understanding System), and the latest version of system, MPLUS (M++). The system was specifically designed for processing chest radiograph reports.

*Background knowledge*

- SPECIALIST lexicon from UMLS, a synonyms database, part-of-speech lexicon.
- An ATN (Augmented Transition Network) grammar, a transformational rule base, and a set of resolution strategy rules.
- Knowledge bases also contain belief network node structures, values, and training cases for each context. The context was pre-defined such as events in chest radiology reports.

*Pipeline*

SymText consists of three primary modules for the analysis and interpretation of sentences [27].
- First, a structural analyzer generates an initial structural interpretation of a sentence.
- Second, a transformational module transforms the initial structure according to the targeted semantic contexts.
- Third, a resolution module semantically resolves the conceptualizations of the text according to its structure. Encoded data are the system's outputs.

SymText's outputs contain three semantic concepts: finding, disease, and appliances (devices).

The distinct feature of SymText when compared to other systems is that it uses belief networks to represent biomedical domain knowledge and discover relationships between nodes within parse trees. SymText has been used in several applications such as mapping chief complaints into ICD-9 codes [29] and extracting pneumonia-related findings from chest radiograph reports [30, 31].

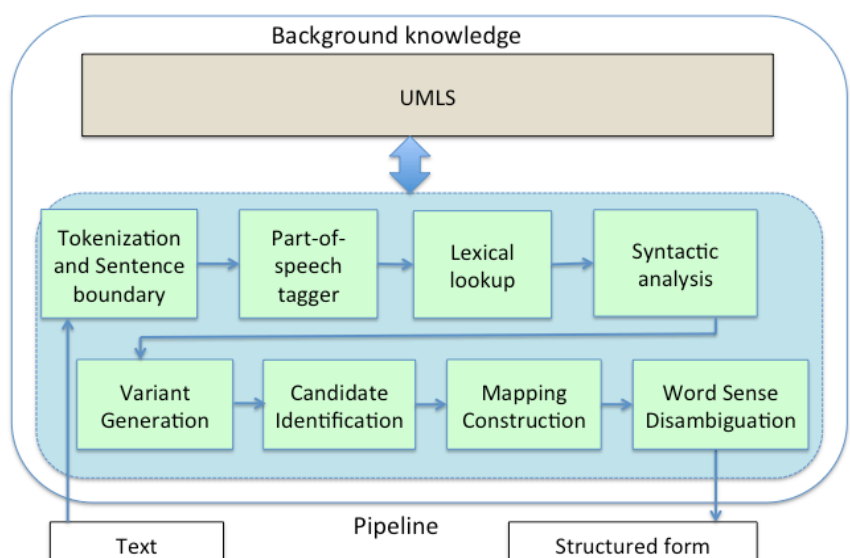

**Fig. 4.** Architecture of the MetaMap system, modified from the original [33], where background knowledge is based on UMLS and different modules represent the pipeline.

Evaluation using 292 chest radiograph reports to identify pneumonia-related concepts showed that the system achieved 0.94 Recall, 0.78 Precision, and 0.84 Specificity, outperforming lay persons [31]. MPLUS was evaluated for the extraction of American College of Radiology utilization review codes from 600 head CT reports. The system achieved 0.87 Recall, 0.98 Specificity and 0.85 Precision in identify reports as positive, i.e., containing brain findings [28].

*MetaMap*

MetaMap[19] [32, 33] was originally developed in 1994 by Aronson at the National Library of Medicine. It was created for mapping the biomedical literature to concepts in the Unified Medical Language System Metathesaurus (UMLS) [2]. It has been widely used for processing clinical text [34–36]. The tool uses a variety of linguistic processes to map from text to Concept Unique Identifiers (CUI) in the UMLS. It is written in Perl, C, Java and Prolog. The architecture of Metamap is depicted in Fig. 4.

---

[19] http://metamap.nlm.nih.gov/

*Background knowledge*

The UMLS is used as the knowledge resource.

*Pipeline*

The most recent version of the system, as described by Aronson and Lang [33], has a two stage architecture:

- Lexical/syntactic processing.
    o Tokenization (including sentence splitting and acronym expansion)
    o Part-of-Speech tagging
    o Lexical lookup that uses the UMLS SPECIALIST lexicon
    o Syntactic analysis that generates phrases for further processing
- Phrasal processing.
    o A table lookup is used to identify variants of phrase words
    o Candidate identification identifies and ranks strings from the UMLS that match phrasal terms
    o Mapping to text through selection, combination and mapping of candidates to the text
    o Word sense disambiguation selects senses consistent with the surrounding text

MetaMap's output can be provided either in XML format, MetaMap Output (MMO), or human readable (HR) formats. Since its initial development MetaMap has been used in a variety of clinical text processing tasks. For example, Shah et al [34] used it to extract cause of death from electronic health records, while Meystre et al. [35] used it to extract medication information from the clinical record. Pakhomov et al. [36] used MetaMap to extract Health Related Quality of Life Indicators from diabetes patients described in physician notes. Recently, Doan et al. [37] used MetaMap for phenotype mapping in the PhenDisco system, a new information retrieval system for the National Center for Biotechnology Information's database of genotypes and phenotypes (dbGaP)[20].

The MetaMap tool is highly configurable, consisting of such advanced features such as *negation detection* (using the NegEx algorithm described in Chapman et al. [38]) and *word sense disambiguation*. Although not open source, the software is freely available for free from the National Library of Medicine as a standalone command-line tool implemented primarily in Prolog.

---

[20] http://www.ncbi.nlm.nih.gov/gap

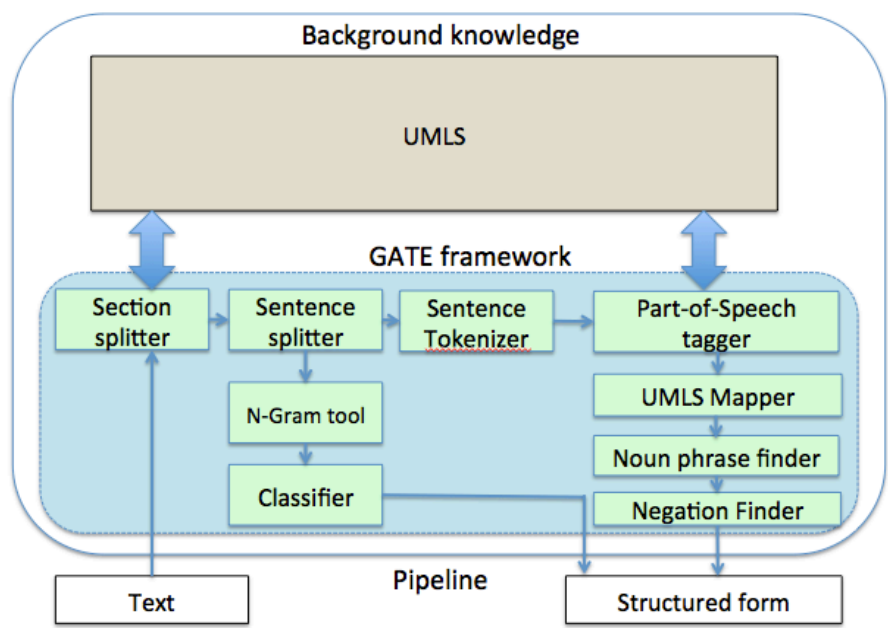

**Fig. 5.** Architecture of HITEx system, simplified from the original publication by Zeng et al [41].

In addition to the Prolog version of MetaMap, a web-based interface is available that facilitates simple queries and also batch processing of text. Furthermore, a Java implementation of MetaMap, MMTx, is available although this version is no longer under active development.

MetaMap was used by the NLM team in the 2009 i2b2 challenge on medication extraction. It achieved an F-score of 0.803, with Precision 0.784 and Recall 0.823. Although it ranked fourth in the challenge, it had the highest Recall among participating teams [39, 40]. Another system that used MetaMap, Textrator, developed by Meystre et al. was also among the top ten in that competition [35, 40].

*HITEx*

HITEx[21] (Health Information Text Extraction) is an open-source NLP system (under i2b2 software license) developed at Brigham and Women's Hospital and Harvard Medical School. It was built based on the GATE framework. The system leverages a set of NLP modules known as CREOLE (a Collection of REusable Objects for Language Engineering) in GATE for low-level processing, such as sentence splitting and part-of-speech tagging. Other components for high-level processor, such as a UMLS mapper and classifier, were developed as plug-in components and are easily handled for loading/reloading. The architecture of HITEx is depicted in Fig. 5.

---

[21] https://www.i2b2.org/software/projects/hitex/hitex_manual.html

*Background knowledge*

HITEx uses UMLS for background knowledge. It has trained corpora for several tasks such as for building a classifier for status-smoking status.

*Pipeline*

HITEx contains following modules integrated in the GATE framework.
- The *Section splitter/filter* splits clinical reports into sections and assigns them to section headers. There are over 1000 section headers in HITEx. Then it filters sections based on selection criteria such as section names, etc.
- The *Sentence splitter* breaks sections into sentences. It is based on regular-based rules.
- The *Sentence tokenizer* breaks sentences into words, it uses an extensive set of regular expressions that define both token delimiters and special cases.
- The *Part-of-speech tagger* assigns part-of-speech (POS) tags to each token in the sentence. This module is a rule-based POS tagger as a plug-in for the GATE framework.
- The *Noun phrase finder* groups POS-tagged words into the noun phrases using the set of rules and the lexicon. This module is a plug-in for the GATE framework.
- The *UMLS mapper* associates the strings of text to UMLS (Unified Medical Language System) concepts. It uses a UMLS dictionary look-up: it first attempts to find exact matches, and when exact matches are not found it stems, normalizes, and truncates the string.
- The *Negation finder* assigns the negation modifier to existing UMLS concepts. It used the NegEx algorithm [38].
- The *N-Gram tool* extracts n-word text fragments along with their frequency from a collection of text.
- The *Classifier* takes a smoking-related sentence to determine the smoking status of a patient. It determines one of following classes: 'current smoker', 'never smoked', 'denies smoking', 'past smoker' or 'not mentioned'.

The system has been used for the extraction of family history from 150 discharge summaries, with accuracies of 0.82 for principal diagnosis, 0.87 for co-morbidity, and 0.90 for smoking status extraction, when excluding cases labeled "Insufficient Data" in the gold standard [41, 42].

*cTAKES*

The cTAKES[22] (clinical Text Analysis and Knowledge Extraction System) system [43], initiated by a Mayo-IBM collaboration in 2000, was first released as an open source toolkit in 2009 by Savova et al. It is an open-source software system under the Apache v2.0 license and is widely used by multiple institutions. The system leverages NLP tools from OpenNLP [44] with trained clinical data from Mayo Clinic. It is the first clinical NLP system to adopt UIMA as its framework.

*Background knowledge*

cTAKES used trained corpora from Mayo clinic data and other sources, utilizing the UMLS as the main background knowledge. Trained corpora were used for low-level processing such as sentence splitting and tokenizing. The UMLS was used for named entity recognition (NER) look-up.

*Pipeline*

cTAKES employs a number of rule-based and machine learning methods. The system can take inputs in plain text or in XML format. It initially included these basic components:

- The *Sentence boundary detector* extends OpenNLP's supervised maximum entropy sentence detection tool.
- The *Tokenizer* breaks sentences into tokens and applies rules to create tokens for date, time, fraction, measurement, person title, range, and roman numerals.
- The *Normalizer* maps multiple mentions of the same word that do not have the same string in the input data. It leverages the SPECIALIST NLP tools[23] from the National of Library Medicine.
- The *POS tagger* and the *Shallow parser* are wrappers around OpenNLP's modules.
- The *Named Entity Recognizer* (NER) uses a dictionary lookup based on noun phrase matching. The dictionary resource is from UMLS. It maps words into UMLS semantic types including diseases/disorders, signs/symptoms, procedure, anatomy and medications. After being mapped into semantic types, name entities are also mapped into UMLS's CUIs.

---

[22] http://ctakes.apache.org/
[23] http://www.specialist.nlm.nih.gov/

cTAKES incorporates the NegEx algorithm [38] for detecting negation from clinical text. Since UIMA is a framework that can easily adapt to new modules, cTAKES integrates other modules such as an assertion module, a dependency parser, a constituency parser, a semantic role labeller, a co-reference resolver, a relation extractor, and a smoker status classifier.

There has been considerable focus on the evaluation of cTAKES core preprocessing modules. The sentence boundary detector achieved an accuracy of 0.949, while tokenizer accuracy was also very high at 0.949. Both part-of-speech tagger and shallow parsing performed well, achieving accuracies of 0.936 and 0.924, respectively. For NER, the system achieved a 0.715 F-score for exact and a 0.824 F-score for overlapping span [43].

cTAKES was first applied to phenotype extraction studies [43] and then was extended to identify document-level patient smoking status [45] and patient level summarization in the first i2b2 challenge [46]. The system was used to generate features for a state-of-the-art system in the 2010 i2b2 challenge on relation extraction of medical problems, tests, and treatments [47].

## Discussion and Conclusion

We provided an overview of several clinical NLP systems under a unified architectural view. Background knowledge plays a crucial role in any clinical NLP task, and currently the UMLS is a major background knowledge component of most systems. Rule-based approaches utilizing the UMLS are still dominant in many clinical NLP systems. Rule-based NLP systems have historically achieved very good performance within specific domains and document types such as radiology reports and discharge summaries. One of the main reasons for using a rule-based approach is that rules are relatively easy to customize and adapt to new domains as well as to different types of clinical text.

Earlier NLP systems such as LSP-MLP and MedLEE are comprised of "hard coded" system modules that do not facilitate reuse. The development of general frameworks such as GATE and UIMA allows sub-tasks or modules to be developed independently and integrated easily into the framework. Machine learning algorithms have been shown to significantly benefit NLP sub-tasks such as NER. Therefore, they can serve as independent modules to be integrated into a framework to improve a sub-task in a clinical NLP system. The combination of machine learning and rule-based approaches in a single hybrid NLP system often achieves better performance than systems based on a single approach. In recent years, a clear trend has developed towards creating reusable NLP modules within open source frameworks like GATE and UIMA.

The main limitation of machine learning when compared to rule-based approaches is that rule-based systems do not require significant amounts of expensive, manually annotated training data, machine learning algorithms typically do. This problem is exacerbated in the biomedical domain, where suitably qualified annotators can be both hard to find and prohibitively expensive [48, 49].

There is an increasing trend towards building community-wide resources and tools for clinical NLP. There have been several shared tasks that bring researchers in clinical NLP together to solve, evaluate and compare different methods. Additionally, there are shared computing resources that aggregate several NLP tools to facilitate the work of researchers, such as the NLP environment in iDASH [50]. The Online Registry of Biomedical Informatics Tools (ORBIT)[24] project is another platform for sharing and collaborating for biomedical researchers in order to create and maintain a software registry, in addition to knowledge bases and data sets. Applications that benefit from biomedical NLP systems, such as EMR linking to genomic information [51], are likely to have great utilization in the next few years. We presented here a unified overview of a few exemplary NLP systems from the architectural perspective that all these systems have two important components: background knowledge and a computational framework. How these components are constructed and integrated into pipelines for biomedical NLP is a critical determinant for their performance.

## Acknowledgements

SD and LOM were funded in part by NIH grants U54HL108460 and UH3HL108785.

---

[24] http://orbit.nlm.nih.gov